\documentclass[11pt,a4paper]{article}
\usepackage[final]{neurips_2022}

\usepackage{times}
\usepackage{latexsym}
\usepackage{graphicx}
\usepackage[T1]{fontenc}
\usepackage[utf8]{inputenc}

\usepackage{booktabs}

\usepackage{microtype}

\usepackage{amsmath,amssymb,amsthm,amscd,amstext}

\newcommand{\CTA}{\texttt{CTA}}
\newcommand{\ATA}{\texttt{ATA}}

\newcommand{\LTA}{\texttt{LTA}}

\newcommand{\beq}{\begin{equation}}
\newcommand{\eeq}{\end{equation}}
\newcommand{\beqs}{\begin{eqnarray}}
\newcommand{\eeqs}{\end{eqnarray}}
\newcommand{\barr}{\begin{array}}
\newcommand{\earr}{\end{array}}

\title{Weakly Supervised Data Augmentation Through Prompting for Dialogue Understanding}

\author{Maximillian Chen$^1$\thanks{Work done during internship at Amazon Alexa AI}~, 
  Alexandros Papangelis$^{2}$, 
  Chenyang Tao$^{2}$,
  Andy Rosenbaum$^2$, \\
  \textbf{Seokhwan Kim}$^2$, 
  \textbf{Yang Liu}$^2$, 
  \textbf{Zhou Yu}$^1$
  \textbf{Dilek Hakkani-Tur}$^2$,\\
  $^1$Columbia University, $^2$Amazon Alexa AI \\
  \texttt{maxchen@cs.columbia.edu}, \texttt{zy2461@columbia.edu} \\
  \texttt{\{papangea, chenyt, andros, seokhwk, yangliud, hakkanit\}@amazon.com} \\
}
\date{}

\begin{document}
\maketitle
\newcommand{\pipeline}{\textsc{WeakDAP }}
\newcommand{\pipelinenospace}{\textsc{WeakDAP}}
\newcommand{\fbtod}{\textsc{FBTOD }}
\newcommand{\fbtodnospace}{\textsc{FBTOD}}
\newcommand{\fbtodfull}{\textsc{Facebook Multilingual Task-Oriented Dialogue }}
\newcommand{\fbtodfullnospace}{\textsc{Facebook Multilingual Task-Oriented Dialogue}}
\newcommand{\dyda}{\textsc{DailyDialog }}
\newcommand{\dydanospace}{\textsc{DailyDialog}}

\begin{abstract}
Dialogue understanding tasks often necessitate abundant annotated data to achieve good performance and that presents challenges in low-resource settings.
To alleviate this barrier, we explore few-shot data augmentation for dialogue understanding by prompting large pre-trained language models and present a novel approach that iterates on augmentation quality by applying weakly-supervised filters.
We evaluate our methods on the emotion and act classification tasks in \dyda and the intent classification task in \fbtodfullnospace. Models fine-tuned on our augmented data mixed with few-shot ground truth data are able to approach or surpass existing full-shot state-of-the-art performance on both datasets. For \dyda specifically, using 10\% of the ground truth data we outperform the current state-of-the-art model which uses 100\% of the data.

\end{abstract}

\section{Introduction \& Related Work}
\setcounter{footnote}{0}
Most common ways of automatic data augmentation in natural language tasks include simple perturbations~\citep{wei-zou-2019-eda, karimi-etal-2021-aeda-easier, xie2020unsupervised} and generative approaches~\citep{kim2021linda,sahu2022data,edunov2018understanding}. However, these methods do not utilize intersentential context, which is essential to encode
for both dialogue understanding and generation. 

On the other hand, modern  pre-trained language models (PLMs) can be prompted to complete dialogues using prefix prompts~\citep{liu2021pre}, which naturally encode conversational context. PLMs also have shown impressive zero- and few-shot capabilities~\citep{brown2020language, bommasani2021opportunities} in dialogue tasks and have been successfully used in generative augmentation frameworks for tasks such as intent classification~\citep{sahu2022data, limaking}, commonsense reasoning~\citep{yang-etal-2020-generative}, and response generation~\citep{kulhanek2021augpt, gao2020paraphrase}. Several studies examine in-context learning, which involves including training examples as part of a prompt~\citep{wei2022chain,min2022rethinking,chen2022meta,lu2022fantastically}. In this work, we take the first step towards applying few-shot prompting to augmenting dialogue datasets. We focus on low-resource\footnote{both in terms of data and cost of computational resources.} settings, contributing an empirical account of augmenting turn-level dialogue understanding tasks using discrete prompting which encodes dialogue history as in-context examples.

\begin{figure}[t]
    \centering
    \includegraphics[width=0.7\linewidth]{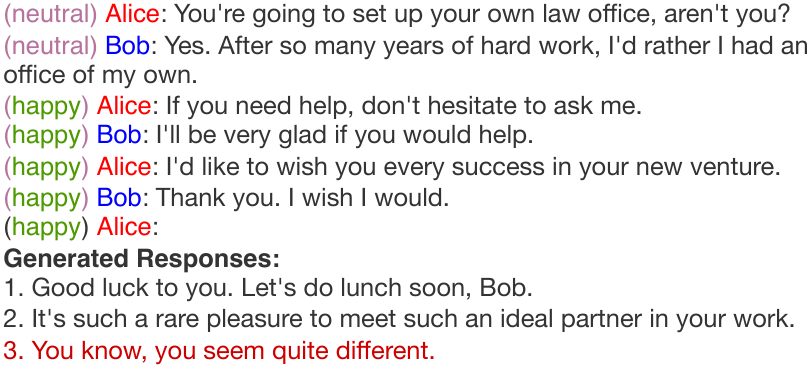}
    \caption{Example augmented conversation from \dyda with a generated turn following the desired emotion ``happy.'' \pipeline filters out generated turns which do not follow the label (red).}
    \label{fig:Figure1}
\end{figure}

One challenge with zero- and few-shot prompting with PLMs is that the outputs may 
exhibit more diversity than one would expect for a specific task, which confounds model training
~\citep{perez2021true, zhao2021calibrate}. 
Specifically, PLMs often synthesize data points which lie outside of the data manifold\footnote{\citet{kim2021linda} hypothesizes that synthetic data must lie along the same natural language manifold as the ground truth data, proposing linear interpolation among existing data.} of a given task, instead following the distribution of the generic pretraining corpora. Due to their distance from the target task's distribution, these augmented samples may be considered low quality. 
We thus propose \pipeline (Weakly supervised Data Augmentation through Prompting), a framework that iteratively improves the quality of augmented data in dialogue classification tasks by introducing a weakly supervised labeler to filter prospective data points. Figure~\ref{fig:Figure1} demonstrates \pipeline filtering out a low-quality synthetic utterance. We demonstrate the effectiveness of \pipeline on emotion and dialogue act classification in \dyda~\citep{li2017dailydialog}, showing on-par or better performance compared to state-of-the-art full-shot results by augmenting only 10\% of the original data. We additionally examine the robustness of \pipeline using a separate task: cross-lingual augmentation for Spanish intent detection in \fbtod~\citep{schuster2019cross}.

\section{Data Augmentation Methods}
Our approach consists of two parts: prompting PLMs using dialogue context, and applying weak supervision to refine prompt-augmented datasets.

\subsection{Constructing Dialogue Prompts}
Dialogue contexts can be used to form prefix prompts which serve as the input to a PLM\footnote{While augmentation by prompting PLMs can help expand linguistic diversity, it can also introduce biases which exist in PLMs' pre-training corpora. Additionally, it may underline biases in the existing low-resource data being augmented. We discuss this further in Appendix~\ref{ethics}.}. We augment the data by replacing dialogue turns, which are selected using the dialogue context construction strategies below. We illustrate specific examples of each in Figure \ref{fig:last_turn_example} and Section \ref{sec:example_prompts} in the Appendix. Each generated utterance can be prescribed a randomly sampled or ground truth reference label. \\

{\bf Conversation Trajectory Augmentation (\CTA).} We take each speaker's first turn as ground-truth context and iteratively replace the next turn with a generated utterance. We autoregressively use each generated utterance as context to generate the next turn. Each ground truth conversation results in one synthetic conversation with a new ``trajectory.'' \\
{\bf All-Turn Augmentation (\ATA).} \ATA~iteratively replaces each turn in the conversation with a generated utterance, but uses the ground truth context instead of the generated context. For a conversation with $n$ turns, this results in $n-1$ ``new'' conversations of length $2$ through $n$.\\
{\bf Last-Turn Augmentation (\LTA).} This is a special case of \ATA~where we simply choose the last turn of the conversation to replace with a generated utterance. This results in the largest conversational context, helping guide the conditional output closer to the ground truth language manifold. Relative to a ground-truth conversation, this yields one new conversation, with an alternate last turn. Example in Figure~\ref{fig:Figure1}.

% {\bf Conversation Trajectory Augmentation (\CTA).} Taking the first turn from each speaker in the data as context, we iteratively replace the next turn with a generated utterance, and autoregressively use the generated utterances as context for the next turn. \\
% {\bf All-Turn Augmentation (\ATA).} \ATA~iteratively replaces each turn in the conversation with a generated utterance, but uses the ground truth context instead of the generated context. \\
% {\bf Last-Turn Augmentation (\LTA).} This is a special case of \ATA~where we simply choose the last turn of the conversation to replace with a generated utterance. This results in the largest conversational context, helping guide the conditional output closer to the ground truth language manifold. 

\subsection{Augmentation with Weak Supervision}

While prompting large PLMs provides a convenient, powerful way to bridge the gap between inadequate training data and data-hungry conversational models, there is a caveat: those PLMs are trained on generic corpora ({\it i.e.}, web crawls, books, {\it etc}.), whose distribution may considerably differ from the data needed to train task-specific models ({\it e.g.}, see Figure \ref{fig:manifold_comparison}). 
This motivates {\it post-hoc} adjustments to make our prompted augmentations more task-aware. 
Weak supervision has been proposed for finding a ``useful representation'' for a task~\citep{robinson2020strength}.
Intuitively, naive prompted augmentations are less potent because they lack task-knowledge\footnote{PLMs only see prompts during generation; to fully account for task knowledge one should include all available examples in-context, which is generally impractical.}, which can be distilled from ground-truth (``gold'') samples by training an auxiliary model. We can then use that model to filter out inconsistent generated utterances. 
\begin{figure*}[t]
    \centering
    \includegraphics[width=\linewidth]{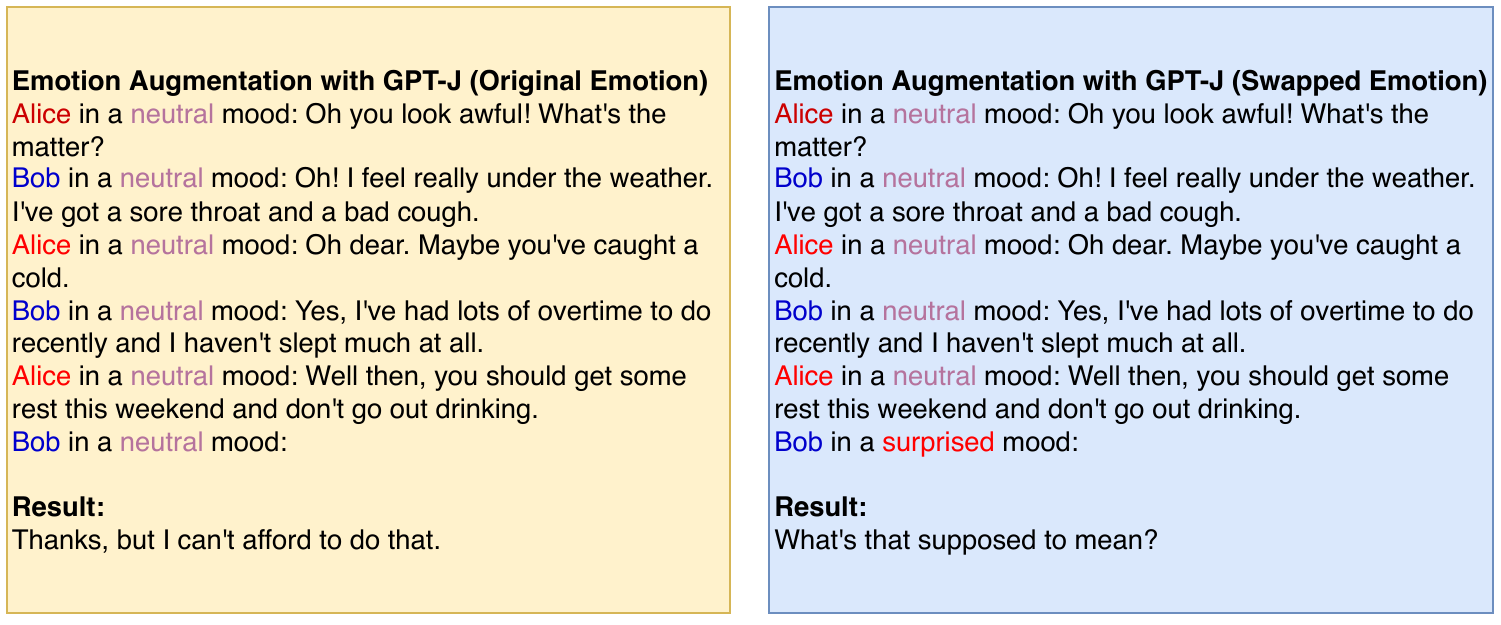}
    \caption{Example conversation augmentation prompt for emotion classification using GPT-J, prescribing the original ground-truth emotion (left) and a randomly sampled emotion (right). This is augmented using Last Turn Augmentation, i.e., the first five turns are taken from the ground-truth data and the model is asked to generate the sixth and final turn. Both boxes represent a new augmented conversation when taken in aggregate.}
    \label{fig:last_turn_example}
\end{figure*}

\begin{figure}[t]
    \centering
    \scalebox{0.60}{
    \includegraphics[width=\linewidth]{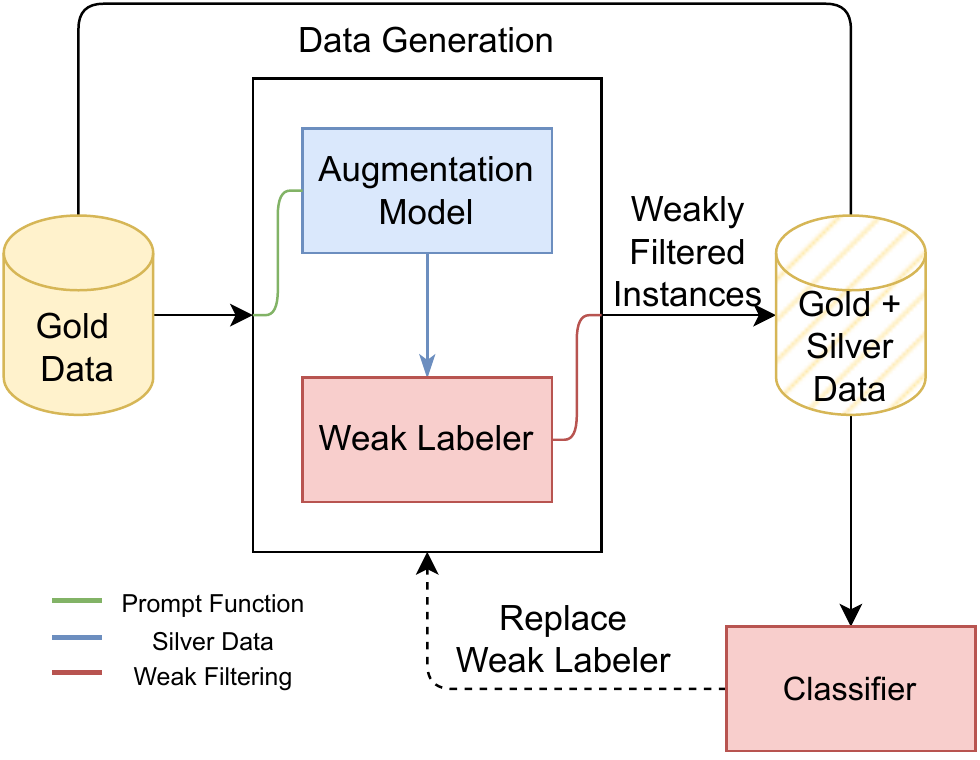}}
    \caption{
    The workflow of \pipelinenospace.
    On each iteration, the Gold Data is augmented by replacing conversation turns generated by providing a PLM with prefix prompts. Each prospective silver training instance is weakly classified as either following its intended label or not, using a task specific classifier. The gold and silver data are used as training data for the next generation's classifier. This process repeats until the performance of the classifier does not improve past a  threshold.}
    \label{fig:pipeline}
\end{figure}

We propose \pipelinenospace, a framework generalizeable to any prompt-based augmentation task. In this work, we prompt GPT-J 6B~\citep{gpt-j} and the Alexa Teacher Model (ATM) 20B~\citep{soltan2022alexatm}. As Figure~\ref{fig:pipeline} illustrates, \pipeline consists of three parts. We first augment the ``gold'' data and train a task classifier on the gold and ``silver'' data. Then, we iteratively re-augment the data and re-train the classifier. For the augmentation step on each iteration, we use the classifier trained during the previous iteration to create a weak silver label for each generated instance, and filter out instances where the silver label does not match the prescribed label with high confidence, i.e., low entropy. We reason that data points which a weak labeler thinks are labeled incorrectly with low confidence could still be useful for learning during training (further discussion in Section \ref{sec:entropy_discussion} in the Appendix). Moreover, this indicates that their labels may be in fact be correct. To this end, we filter out incorrect instances classified in the bottom 80th percentile of entropy, computed as in the equation below, where C is the number of classes and $p_i$ is the probability of class $i$.\footnote{This threshold is tunable.}
\begin{center}
\vspace{-10mm}
\beq
$$\[Entropy = \sum_{i}^{C} -p_i * log_2(p_i)\]$$
\eeq
\end{center}
% \vspace{-5mm}

This weakly guarantees that the generated data is not of low-quality. This continues until the classifier's performance doesn't improve by at least $\epsilon$ for $k$ rounds. Here, we fix $\epsilon=0.005, k=3$. 
\paragraph{Other Task-Aware Augmentation Approaches.}
Similar task-aware generative augmentation approaches typically distill task-knowledge into the generator.
\citet{yang-etal-2020-generative} proposes augmentation for commonsense reasoning by fine-tuning two generators (for answering and distracting) and relabelling synthetic data points using a task model, while \citet{papangelis-etal-2021-generative} fine-tunes a generator using reinforcement learning.
With large PLMs, these methods are costly and less practical.
While few-shot prompting is a cheaper solution, it is less effective at encoding lots of task knowledge, as in-context example capacity is limited.
\pipeline bridges the gap between prompt-based augmentation with little task-knowledge and complex mechanisms with higher computational costs; it does not need to fine-tune the generator, as we prompt it using dialogue context as in-context utterance examples. 
\section{Experiments}
We benchmark various augmentation methods on the classification tasks in \dydanospace, a high-quality open-domain dialogue dataset, and the intent detection task of \fbtodnospace, a task-oriented dialogue dataset (dataset details in Figure~\ref{sec:datasets}). 
%, and our code will be available from \url{https://github.com/author_name/WeakDAP}.

\subsection{\dyda Emotion Classification}
We first conduct a thorough evaluation of our augmentation methods using the emotion classification task in \dyda as a case study, in the full and few-shot settings\footnote{We randomly sample $1\%$, $5\%$, and $10\%$ of the data.}.  For our augmentation model, we use GPT-J 6B\footnote{We examined OPT-30B~\citep{zhang2022opt}, but it was far slower without large performance improvements.}~\citep{gpt-j}, which is one of the largest causal language models publicly available, and has been able to achieve performance competitive to GPT-3 on many tasks~\citep{mesh-transformer-jax,black2022gpt}. For all \dyda experiments we use the Speaker Turn Model (STM) ~\citep{he2021speaker}, a RoBERTa~\citep{liu2019roberta}-based classification model with speaker turn awareness\footnote{STM achieves state-of-the-art performance on full-shot \dyda act classification ($87.5\%$ accuracy).}, as the classification task model and weak labeler.

There are seven emotion labels: {\it neutral, anger, disgust, fear, happiness, sadness}, and {\it surprise}. Each label is a rich, descriptive token on its own, so in constructing a prompt, we directly use it as an adjective (e.g., ``Alice in a {\it happy} mood:''). Additionally, 
we conjecture that directly using conversation history forms the best set of in-context examples to generate utterances which convey a prescribed emotion while remaining within the gold data manifold.
Example prompts provided in Figure \ref{fig:last_turn_example} (\LTA) and Section \ref{sec:example_prompts} in the Appendix (\ATA, \CTA). In Figure~\ref{fig:manifold_comparison}, we compare the data manifold of the synthetic data resulting from \LTA~and \CTA~with a maximum augmentation size of 2x that of the amount of original data used, and \ATA~which results in a size between $5.7$x and $6.2$x (see Appendix Table~\ref{allturn_classification}). We can see that \CTA~results in a separate cluster of data, likely due to the underlying distribution of GPT-J's pretraining data. \ATA's distribution has a clear distinct cluster as with \CTA, but also posseses some overlap with the training distribution. In contrast, we see that synthetic data from \LTA~lies within the training data manifold. This is likely due to the in-distribution context provided as conditional input to GPT-J. We hypothesize that this context most closely guides synthetic data towards the original data distribution. This may be more beneficial in tasks where we do not expect distribution shift from training to inference. In tasks such as response generation, where diverse output is desirable, it may be more beneficial to have training data that falls into an expanded manifold as with \CTA~or \ATA. Comparing the three methods, we find that \CTA~and \ATA~(Appendix Figure \ref{fig:emotion_trajectory}, Table~\ref{allturn_classification}) are beneficial, but underperform \LTA~(Figure \ref{fig:emotion_laststep}) for this task. Thus, we primarily experiment using \LTA.
\begin{figure}
    \centering
    \includegraphics[width=0.5\linewidth]{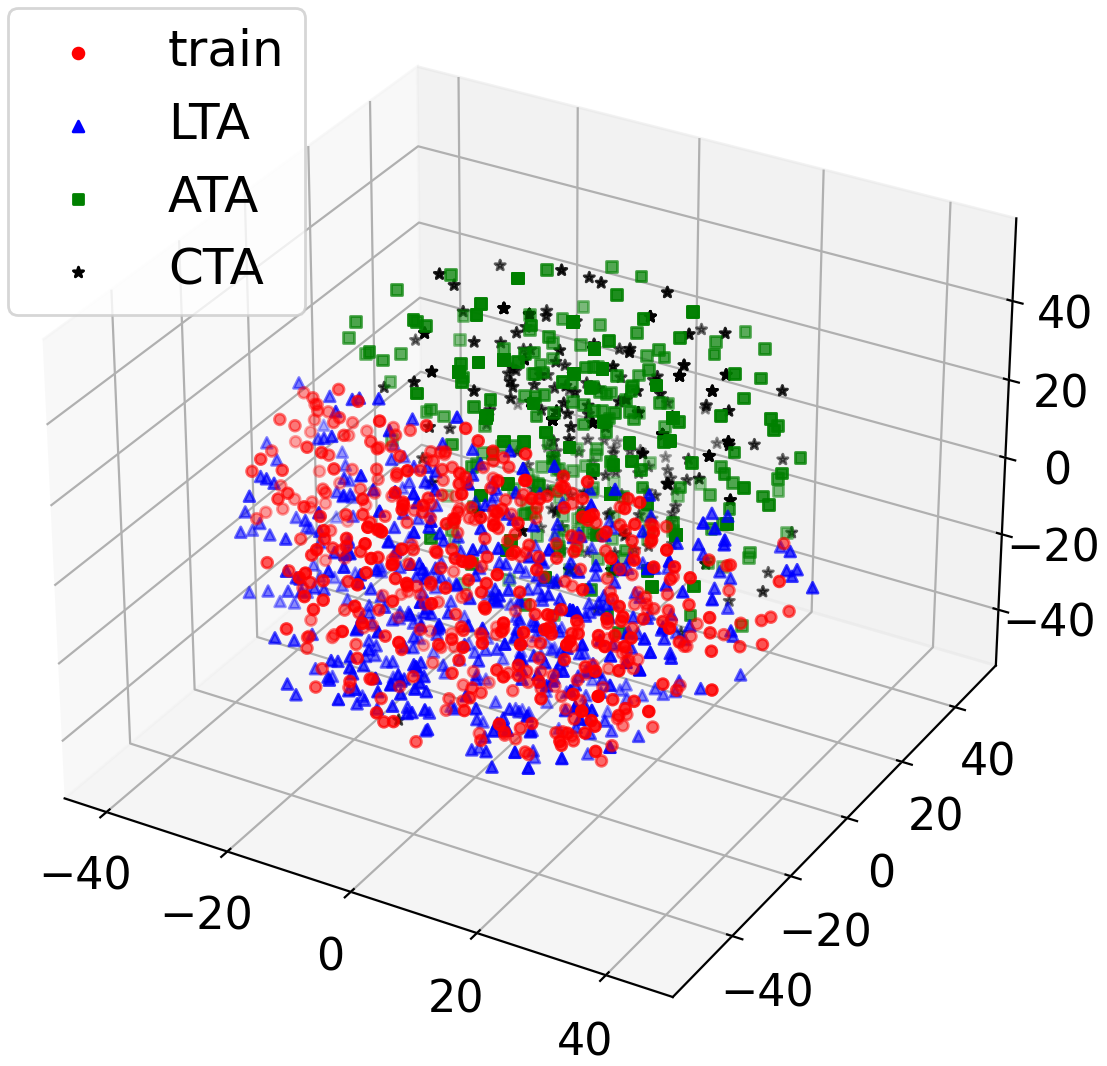}
    \caption{t-SNE projection of a random sample (for clarity) of training and augmented data.}
    \label{fig:manifold_comparison}
\end{figure}

In low-resource settings, it is important to quantify how much ground truth data is available. When augmenting seed data, it is essential to quantify how much data is being generated (i.e., by providing a Size Multiplier relative to the amount of seed data used; \citet{feng-etal-2020-genaug}). Figure \ref{fig:emotion_laststep} shows the performance of STM in each data setting with varying amounts of synthetic data resulting from \LTA. Following existing work, we report micro F1 ignoring the majority label, due to a heavy imbalance towards the neutral emotion~\citep{liang2021s+,lee2021graph,wang2020contextualized}. We see that STM with augmented 10\% data reaches an F1 score of $0.686$ and  $0.70$ with augmented full-shot data, surpassing the existing state-of-the-art of $0.641$ set by S+PAGE~\citep{liang2021s+}.
We observe that adding too much augmented data eventually hurts performance in each setting, further suggesting that prompting alone has inconsistent quality. 2x is the best performing multiplier and we use that to conduct all other experiments. 
Figure \ref{fig:weakdap_benefits} also shows that \pipeline improved STM's performance in each few-shot setting. On average, the labeler reduced the augmented data size from 2x to 1.8x, indicating that it may foster more efficient learning. 

Prior work has not found large differences in performance between backtranslation and other perturbation methods despite higher computational overhead~\citep{xie2020unsupervised, kim2021linda}. Thus, we primarily compare against EDA~\citep{wei-zou-2019-eda}\footnote{EDA includes synonym replacement, random insertion, random swap, and random deletion.} and AEDA~\citep{karimi-etal-2021-aeda-easier}\footnote{AEDA randomly inserts punctuation marks into text.} as noise injection baselines. As seen in Figure \ref{fig:weakdap_benefits}, using noise perturbed data underperforms normal data, an outcome corroborated by \citet{kumar2020data,chen2021empirical}. Finally, we examined the importance of context by experimenting with a standard in-context learning prompt with 10 randomly sampled utterances of the same emotion label~\citep{sahu2022data,brown2020language}, finding that dialogue context consistently outperforms random sampling (e.g. 0.624 versus 0.600 F1 using $5\%$ data).

\begin{figure}
\parbox{.49\linewidth}{
    \centering
    \includegraphics[width=\linewidth]{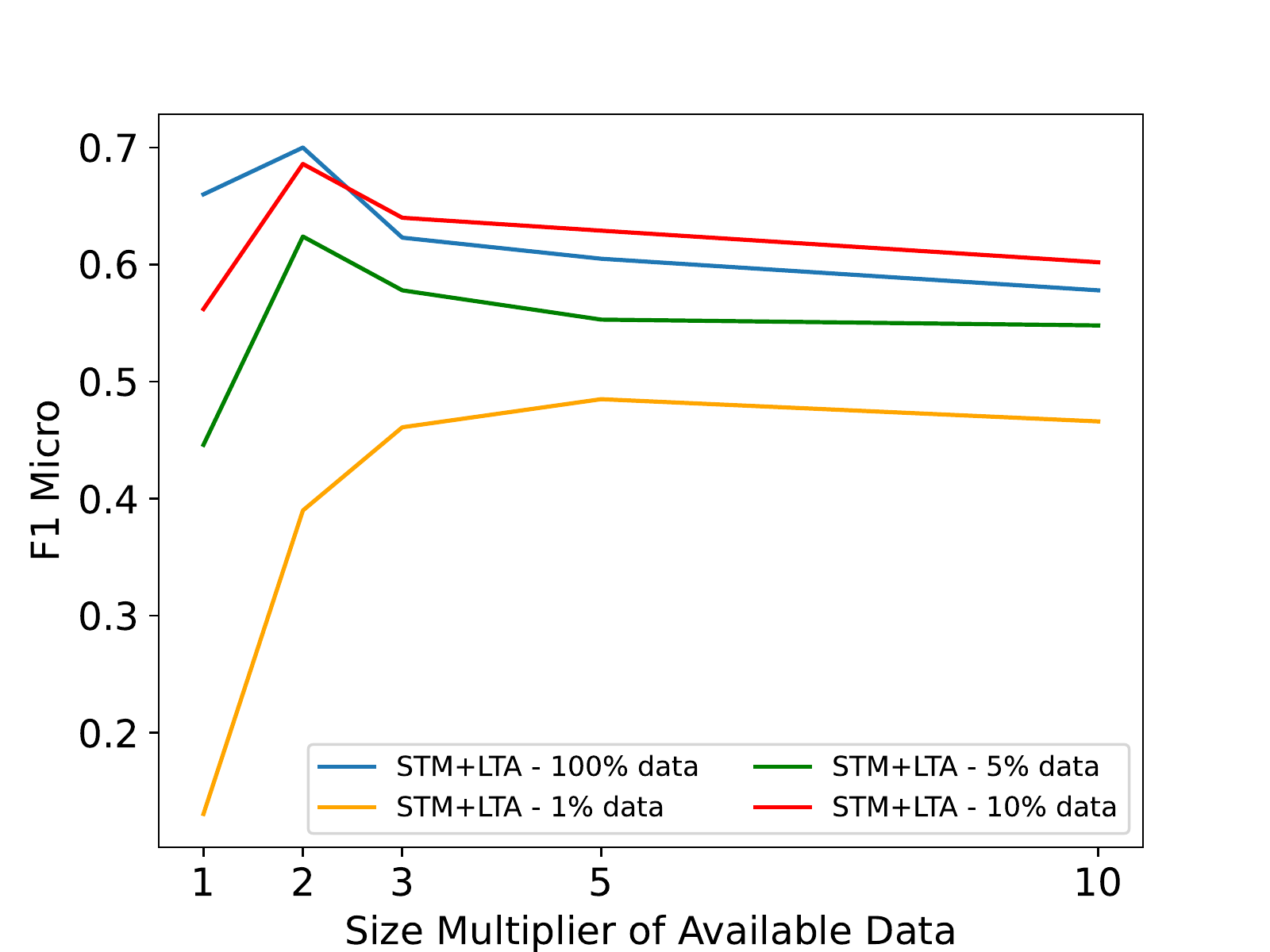}
    \caption{Micro F1 on the \dyda emotion classification task using \LTA~for different data sizes.
    }
    \label{fig:emotion_laststep}
}
\hspace{3mm}
\parbox{.49\linewidth}{
    \vspace{8mm}
    \centering
    \includegraphics[width=\linewidth]{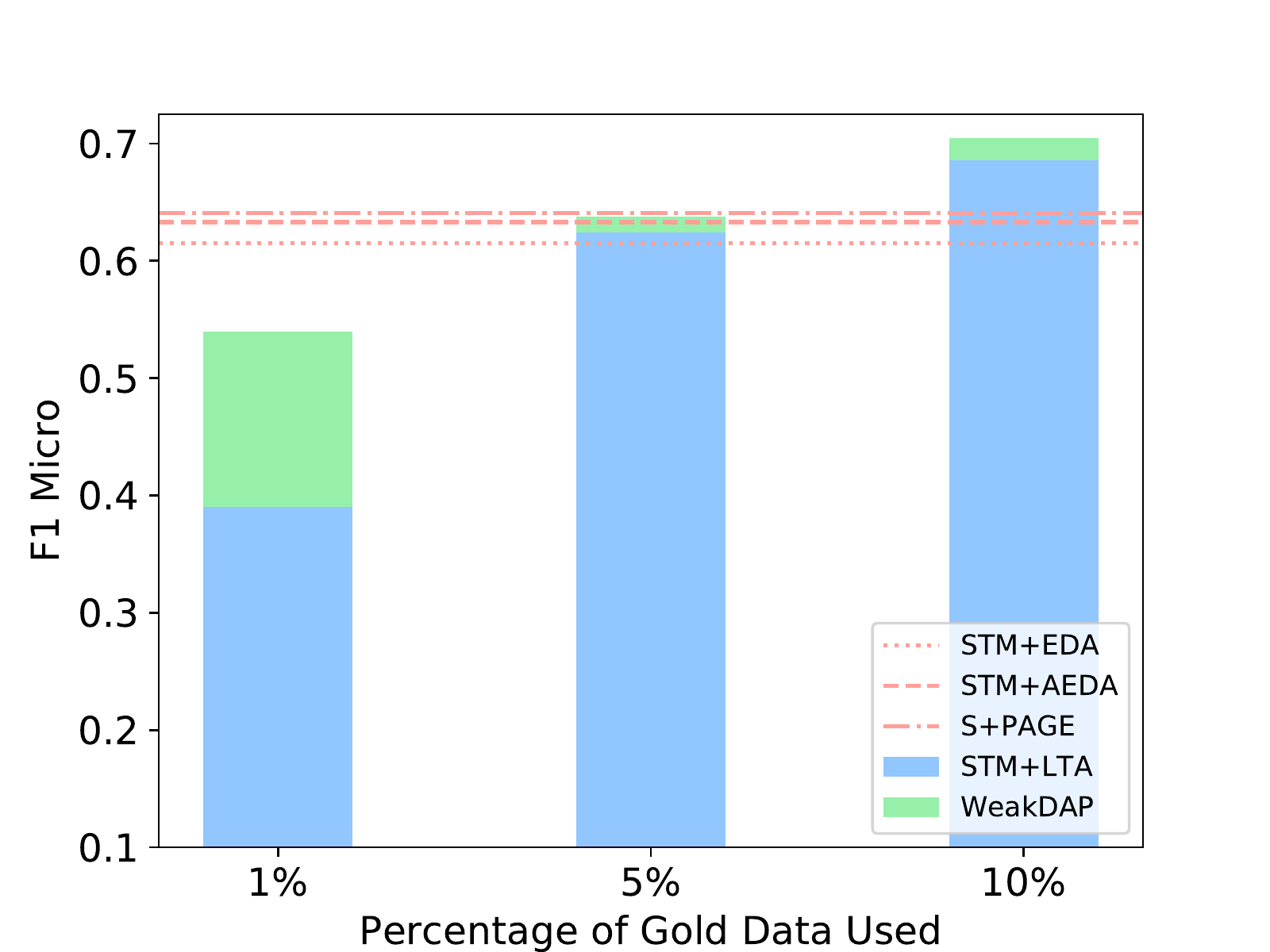}
    \caption{
    Comparison of prompt-augmentation with \LTA~and \pipelinenospace. \pipeline with $10\%$ data outperforms full-shot state-of-the-art performance with S+PAGE, and baseline augmentation approaches mixed with full-shot data.
    }
    \label{fig:weakdap_benefits}
}
\end{figure}

\subsection{\dyda Act Classification}
There are four dialogue act labels: \emph{inform, question, directive}, and \emph{commissive}. While these individual terms are less descriptive than emotion labels, they form descriptive tokens if used as active verbs (e.g., ``Alice directs Bob:''). See Section \ref{sec:example_prompts} in the Appendix for an example.

Following our findings on Emotion Classification, we augment the few-shot \dyda Act Classification task with \LTA~while providing full-shot performance for reference. Our results in Table~\ref{act_classification} indicate that \pipeline results in the highest accuracy across all few-shot settings, although using STM on the unaugmented data achieves the highest F1 in the 5\% setting. Noticeably, the performance improvements on the act classification are not as drastic as in the emotion classification task. This is likely because the baseline performance on the act classification task was already competitive.

\subsection{Cross-lingual Augmentation}
To assess the robustness of \pipelinenospace, we additionally evaluated its performance on a completely different setting: low-resource cross-lingual augmentation for intent classification on \fbtodnospace. For augmentation, we used the Alexa Teacher Model \citep{soltan2022alexatm}, a 20B multi-lingual seq2seq language model pre-trained on a text de-noising objective similar to \citet{liu2020multilingual} using large publicly available datasets including mC4~\citep{xue2021mt5} and Wikipedia. For intent classification, we fine-tuned XLMRoBERTa (XLMR;~\citet{conneau2020unsupervised}).

Full-shot fine-tuning of XLMR yields an accuracy of $98.8\%$ on the Spanish test set, however we focus on the few-shot Spanish setting, with few- and full-shot cross-lingual augmentation from English. For all few-shot datasets, we use $1\%, 5\%,$ and $10\%$ of the original training data for that language (see Section \ref{sec:crosslingualsetup}). \fbtod is a single-turn dataset, so we adapt the in-context learning prompt in
\citet{sahu2022data}, mixing the reference Spanish instance with randomly sampled English instances of the same label for in-context learning (example in Section \ref{sec:example_prompts} in the Appendix). We do not examine Thai because the model has not been pre-trained on Thai data.

We augment the few-shot Spanish setting using both low-resource and high-resource English data.
With low-resource English, we use the same percentage of data as the ground-truth Spanish data. Table~\ref{intent} indicates that fine-tuning XLMR on \pipeline outperforms prompt-based augmentation in all settings both in terms of accuracy and F1 score.

\begin{table}[]
\parbox{.45\linewidth}{
\centering
\begin{tabular}{lll}
\toprule
\dyda        & Accuracy & F1   \\ 
\midrule
1\% No Aug.          & 0.789    & 0.648 \\
1\% Prompt    & 0.792     & 0.714  \\
~~~+ WeakDAP   & \textbf{0.796}     & \textbf{0.716}  \\ \hline
5\% No Aug.        & 0.817     & \textbf{0.809} \\
5\% Prompt    & 0.828     & 0.762 \\
~~~+ WeakDAP   & \textbf{0.832}         & 0.765     \\ \hline
10\% No Aug.         & 0.839         & 0.802 \\
10\% Prompt   & 0.839         & 0.815     \\
~~~+ WeakDAP  & \textbf{0.842}         & \textbf{0.820}     \\ \hline
100\% SotA & 0.875 & ---\\
\bottomrule
\end{tabular}
\caption{Accuracy and Micro F1 (ignoring majority label) on DailyDialog Act classification using STM with few-shot data. We augment using \LTA~resulting in data sizes of at most two times the original data size.}
\label{act_classification}
}\hspace{3mm}
\parbox{.45\linewidth}{
\centering
\begin{tabular}{lllll}
\toprule
\fbtod ES     & Acc$_{LR}$ & F1$_{LR}$ & Acc$_{HR}$ & F1$_{HR}$   \\ 
\midrule
1\% No Aug. & 0.572    & 0.164 & 0.572    & 0.164\\
1\% Prompt  & 0.737     & 0.316 & 0.776 & 0.359 \\
~~~+ WeakDAP & \textbf{0.834}     & \textbf{0.495} & \textbf{0.831} & \textbf{0.528}\\ \hline
5\% No Aug. & 0.845     & 0.417 & 0.845     & 0.417 \\
5\% Prompt  & 0.953 & 0.641 & 0.954     & 0.682 \\
~~~+ WeakDAP & \textbf{0.957} & \textbf{0.715} & \textbf{0.962}         & \textbf{0.732}     \\ \hline
10\% No Aug. & 0.942         &  0.588 & 0.942         &  0.588    \\
10\% Prompt &  0.973 & 0.772 & 0.973        &  0.791    \\
~~~+ WeakDAP & \textbf{0.979} & \textbf{0.905} &  \textbf{0.976}                            &    \textbf{0.846} \\ \hline
100\% No Aug. & 0.988 & 0.889 & 0.988 & 0.889\\
\bottomrule
\end{tabular}

\caption{Accuracy and Macro F1 on the FBTOD Spanish dataset. 
$HR/LR$: High-resource (full-shot) and Low-resource (few-shot matching the Spanish percentage) English Training data.
}
\label{intent}
}
\end{table}
\section{Conclusion}
We contribute significant progress towards few-shot prompt-based augmentation for dialogue tasks. We introduce \pipeline and demonstrate its augmentation quality filtering capabilities by surpassing full-shot state-of-the-art performance with few-shot examples on \dyda and achieving strong few-shot performance on \fbtodnospace. In the future, we will examine ways to integrate soft prompting into \pipeline as well as identify an appropriate feedback mechanism for response generation.

\bibliographystyle{plainnat}
\bibliography{references}

\begin{thebibliography}{55}
\providecommand{\natexlab}[1]{#1}
\providecommand{\url}[1]{\texttt{#1}}
\expandafter\ifx\csname urlstyle\endcsname\relax
  \providecommand{\doi}[1]{doi: #1}\else
  \providecommand{\doi}{doi: \begingroup \urlstyle{rm}\Url}\fi

\bibitem[Barikeri et~al.(2021)Barikeri, Lauscher, Vuli{\'c}, and
  Glava{\v{s}}]{barikeri2021redditbias}
Soumya Barikeri, Anne Lauscher, Ivan Vuli{\'c}, and Goran Glava{\v{s}}.
\newblock Redditbias: A real-world resource for bias evaluation and debiasing
  of conversational language models.
\newblock \emph{arXiv preprint arXiv:2106.03521}, 2021.

\bibitem[Black et~al.(2022)Black, Biderman, Hallahan, Anthony, Gao, Golding,
  He, Leahy, McDonell, Phang, et~al.]{black2022gpt}
Sidney Black, Stella Biderman, Eric Hallahan, Quentin Anthony, Leo Gao,
  Laurence Golding, Horace He, Connor Leahy, Kyle McDonell, Jason Phang, et~al.
\newblock Gpt-neox-20b: An open-source autoregressive language model.
\newblock In \emph{Proceedings of BigScience Episode \# 5--Workshop on
  Challenges \& Perspectives in Creating Large Language Models}, pages 95--136,
  2022.

\bibitem[Bommasani et~al.(2021)Bommasani, Hudson, Adeli, Altman, Arora, von
  Arx, Bernstein, Bohg, Bosselut, Brunskill,
  et~al.]{bommasani2021opportunities}
Rishi Bommasani, Drew~A Hudson, Ehsan Adeli, Russ Altman, Simran Arora, Sydney
  von Arx, Michael~S Bernstein, Jeannette Bohg, Antoine Bosselut, Emma
  Brunskill, et~al.
\newblock On the opportunities and risks of foundation models.
\newblock \emph{arXiv preprint arXiv:2108.07258}, 2021.

\bibitem[Brown et~al.(2020)Brown, Mann, Ryder, Subbiah, Kaplan, Dhariwal,
  Neelakantan, Shyam, Sastry, Askell, et~al.]{brown2020language}
Tom Brown, Benjamin Mann, Nick Ryder, Melanie Subbiah, Jared~D Kaplan, Prafulla
  Dhariwal, Arvind Neelakantan, Pranav Shyam, Girish Sastry, Amanda Askell,
  et~al.
\newblock Language models are few-shot learners.
\newblock \emph{Advances in neural information processing systems},
  33:\penalty0 1877--1901, 2020.

\bibitem[Chen et~al.(2021)Chen, Tam, Raffel, Bansal, and
  Yang]{chen2021empirical}
Jiaao Chen, Derek Tam, Colin Raffel, Mohit Bansal, and Diyi Yang.
\newblock An empirical survey of data augmentation for limited data learning in
  nlp.
\newblock \emph{arXiv preprint arXiv:2106.07499}, 2021.

\bibitem[Chen et~al.(2020)Chen, Zhang, and Lu]{chen2020uncertainty}
Wenshi Chen, Bowen Zhang, and Mingyu Lu.
\newblock Uncertainty quantification for multilabel text classification.
\newblock \emph{Wiley Interdisciplinary Reviews: Data Mining and Knowledge
  Discovery}, 10\penalty0 (6):\penalty0 e1384, 2020.

\bibitem[Chen et~al.(2022)Chen, Zhong, Zha, Karypis, and He]{chen2022meta}
Yanda Chen, Ruiqi Zhong, Sheng Zha, George Karypis, and He~He.
\newblock Meta-learning via language model in-context tuning.
\newblock In \emph{Proceedings of the 60th Annual Meeting of the Association
  for Computational Linguistics (Volume 1: Long Papers)}, pages 719--730, 2022.

\bibitem[Christian(2020)]{christian2020alignment}
Brian Christian.
\newblock \emph{The alignment problem: Machine learning and human values}.
\newblock WW Norton \& Company, 2020.

\bibitem[Conneau et~al.(2020)Conneau, Khandelwal, Goyal, Chaudhary, Wenzek,
  Guzm{\'a}n, Grave, Ott, Zettlemoyer, and Stoyanov]{conneau2020unsupervised}
Alexis Conneau, Kartikay Khandelwal, Naman Goyal, Vishrav Chaudhary, Guillaume
  Wenzek, Francisco Guzm{\'a}n, {\'E}douard Grave, Myle Ott, Luke Zettlemoyer,
  and Veselin Stoyanov.
\newblock Unsupervised cross-lingual representation learning at scale.
\newblock In \emph{Proceedings of the 58th Annual Meeting of the Association
  for Computational Linguistics}, pages 8440--8451, 2020.

\bibitem[Cs{\'a}ky et~al.(2019)Cs{\'a}ky, Purgai, and
  Recski]{csaky2019improving}
Rich{\'a}rd Cs{\'a}ky, Patrik Purgai, and G{\'a}bor Recski.
\newblock Improving neural conversational models with entropy-based data
  filtering.
\newblock In \emph{Proceedings of the 57th Annual Meeting of the Association
  for Computational Linguistics}, pages 5650--5669, 2019.

\bibitem[Edunov et~al.(2018)Edunov, Ott, Auli, and
  Grangier]{edunov2018understanding}
Sergey Edunov, Myle Ott, Michael Auli, and David Grangier.
\newblock Understanding back-translation at scale.
\newblock In \emph{Proceedings of the 2018 Conference on Empirical Methods in
  Natural Language Processing}, pages 489--500, 2018.

\bibitem[Faal et~al.(2022)Faal, Schmitt, and Yu]{faal2022reward}
Farshid Faal, Ketra Schmitt, and Jia~Yuan Yu.
\newblock Reward modeling for mitigating toxicity in transformer-based language
  models.
\newblock \emph{Applied Intelligence}, pages 1--15, 2022.

\bibitem[Feng et~al.(2020)Feng, Gangal, Kang, Mitamura, and
  Hovy]{feng-etal-2020-genaug}
Steven~Y. Feng, Varun Gangal, Dongyeop Kang, Teruko Mitamura, and Eduard Hovy.
\newblock {G}en{A}ug: Data augmentation for finetuning text generators.
\newblock In \emph{Proceedings of Deep Learning Inside Out (DeeLIO): The First
  Workshop on Knowledge Extraction and Integration for Deep Learning
  Architectures}, pages 29--42, Online, November 2020. Association for
  Computational Linguistics.
\newblock \doi{10.18653/v1/2020.deelio-1.4}.
\newblock URL \url{https://aclanthology.org/2020.deelio-1.4}.

\bibitem[Gao et~al.(2020{\natexlab{a}})Gao, Biderman, Black, Golding, Hoppe,
  Foster, Phang, He, Thite, Nabeshima, et~al.]{gao2020pile}
Leo Gao, Stella Biderman, Sid Black, Laurence Golding, Travis Hoppe, Charles
  Foster, Jason Phang, Horace He, Anish Thite, Noa Nabeshima, et~al.
\newblock The pile: An 800gb dataset of diverse text for language modeling.
\newblock \emph{arXiv preprint arXiv:2101.00027}, 2020{\natexlab{a}}.

\bibitem[Gao et~al.(2020{\natexlab{b}})Gao, Zhang, Ou, and
  Yu]{gao2020paraphrase}
Silin Gao, Yichi Zhang, Zhijian Ou, and Zhou Yu.
\newblock Paraphrase augmented task-oriented dialog generation.
\newblock In \emph{Proceedings of the 58th Annual Meeting of the Association
  for Computational Linguistics}, pages 639--649, 2020{\natexlab{b}}.

\bibitem[Guo and Viktor(2004)]{guo2004boosting}
Hongyu Guo and Herna~L Viktor.
\newblock Boosting with data generation: improving the classification of hard
  to learn examples.
\newblock In \emph{International Conference on Industrial, Engineering and
  Other Applications of Applied Intelligent Systems}, pages 1082--1091.
  Springer, 2004.

\bibitem[He et~al.(2021)He, Tavabi, Lerman, and Soleymani]{he2021speaker}
Zihao He, Leili Tavabi, Kristina Lerman, and Mohammad Soleymani.
\newblock Speaker turn modeling for dialogue act classification.
\newblock In \emph{Findings of the Association for Computational Linguistics:
  EMNLP 2021}, pages 2150--2157, 2021.

\bibitem[Karimi et~al.(2021)Karimi, Rossi, and
  Prati]{karimi-etal-2021-aeda-easier}
Akbar Karimi, Leonardo Rossi, and Andrea Prati.
\newblock {AEDA}: An easier data augmentation technique for text
  classification.
\newblock In \emph{Findings of the Association for Computational Linguistics:
  EMNLP 2021}, pages 2748--2754, Punta Cana, Dominican Republic, November 2021.
  Association for Computational Linguistics.
\newblock \doi{10.18653/v1/2021.findings-emnlp.234}.
\newblock URL \url{https://aclanthology.org/2021.findings-emnlp.234}.

\bibitem[Kim et~al.(2021)Kim, Jeong, and Cho]{kim2021linda}
Yekyung Kim, Seohyeong Jeong, and Kyunghyun Cho.
\newblock Linda: Unsupervised learning to interpolate in natural language
  processing.
\newblock \emph{arXiv preprint arXiv:2112.13969}, 2021.

\bibitem[Kulh{\'a}nek et~al.(2021)Kulh{\'a}nek, Hude{\v{c}}ek, Nekvinda, and
  Du{\v{s}}ek]{kulhanek2021augpt}
Jon{\'a}{\v{s}} Kulh{\'a}nek, Vojt{\v{e}}ch Hude{\v{c}}ek, Tom{\'a}{\v{s}}
  Nekvinda, and Ond{\v{r}}ej Du{\v{s}}ek.
\newblock Augpt: Auxiliary tasks and data augmentation for end-to-end dialogue
  with pre-trained language models.
\newblock In \emph{Proceedings of the 3rd Workshop on Natural Language
  Processing for Conversational AI}, pages 198--210, 2021.

\bibitem[Kumar et~al.(2020)Kumar, Choudhary, and Cho]{kumar2020data}
Varun Kumar, Ashutosh Choudhary, and Eunah Cho.
\newblock Data augmentation using pre-trained transformer models.
\newblock In \emph{Proceedings of the 2nd Workshop on Life-long Learning for
  Spoken Language Systems}, pages 18--26, 2020.

\bibitem[Lauscher et~al.(2021)Lauscher, L{\"u}ken, and
  Glava{\v{s}}]{lauscher2021sustainable}
Anne Lauscher, Tobias L{\"u}ken, and Goran Glava{\v{s}}.
\newblock Sustainable modular debiasing of language models.
\newblock \emph{arXiv preprint arXiv:2109.03646}, 2021.

\bibitem[Lee and Choi(2021)]{lee2021graph}
Bongseok Lee and Yong~Suk Choi.
\newblock Graph based network with contextualized representations of turns in
  dialogue.
\newblock In \emph{Proceedings of the 2021 Conference on Empirical Methods in
  Natural Language Processing}, pages 443--455, 2021.

\bibitem[Li et~al.(2017)Li, Su, Shen, Li, Cao, and Niu]{li2017dailydialog}
Yanran Li, Hui Su, Xiaoyu Shen, Wenjie Li, Ziqiang Cao, and Shuzi Niu.
\newblock Dailydialog: A manually labelled multi-turn dialogue dataset.
\newblock In \emph{Proceedings of the Eighth International Joint Conference on
  Natural Language Processing (Volume 1: Long Papers)}, pages 986--995, 2017.

\bibitem[Li et~al.(2021)Li, Wang, Albalak, Yang, Qian, Li, and Yan]{limaking}
Zekun Li, Hong Wang, Alon Albalak, Yingrui Yang, Jing Qian, Shiyang Li, and
  Xifeng Yan.
\newblock Making something out of nothing: Building robust task-oriented
  dialogue systems from scratch.
\newblock In \emph{Proceedings of Alexa Prize TaskBot}, 2021.

\bibitem[Liang et~al.(2021)Liang, Yang, Xu, Huang, Wang, and Dong]{liang2021s+}
Chen Liang, Chong Yang, Jing Xu, Juyang Huang, Yongliang Wang, and Yang Dong.
\newblock S+ page: A speaker and position-aware graph neural network model for
  emotion recognition in conversation.
\newblock \emph{arXiv preprint arXiv:2112.12389}, 2021.

\bibitem[Liu et~al.(2021)Liu, Yuan, Fu, Jiang, Hayashi, and Neubig]{liu2021pre}
Pengfei Liu, Weizhe Yuan, Jinlan Fu, Zhengbao Jiang, Hiroaki Hayashi, and
  Graham Neubig.
\newblock Pre-train, prompt, and predict: A systematic survey of prompting
  methods in natural language processing.
\newblock \emph{arXiv preprint arXiv:2107.13586}, 2021.

\bibitem[Liu et~al.(2019)Liu, Ott, Goyal, Du, Joshi, Chen, Levy, Lewis,
  Zettlemoyer, and Stoyanov]{liu2019roberta}
Yinhan Liu, Myle Ott, Naman Goyal, Jingfei Du, Mandar Joshi, Danqi Chen, Omer
  Levy, Mike Lewis, Luke Zettlemoyer, and Veselin Stoyanov.
\newblock Roberta: A robustly optimized bert pretraining approach.
\newblock \emph{arXiv preprint arXiv:1907.11692}, 2019.

\bibitem[Liu et~al.(2020)Liu, Gu, Goyal, Li, Edunov, Ghazvininejad, Lewis, and
  Zettlemoyer]{liu2020multilingual}
Yinhan Liu, Jiatao Gu, Naman Goyal, Xian Li, Sergey Edunov, Marjan
  Ghazvininejad, Mike Lewis, and Luke Zettlemoyer.
\newblock Multilingual denoising pre-training for neural machine translation.
\newblock \emph{Transactions of the Association for Computational Linguistics},
  8:\penalty0 726--742, 2020.

\bibitem[Lu et~al.(2022)Lu, Bartolo, Moore, Riedel, and
  Stenetorp]{lu2022fantastically}
Yao Lu, Max Bartolo, Alastair Moore, Sebastian Riedel, and Pontus Stenetorp.
\newblock Fantastically ordered prompts and where to find them: Overcoming
  few-shot prompt order sensitivity.
\newblock In \emph{Proceedings of the 60th Annual Meeting of the Association
  for Computational Linguistics (Volume 1: Long Papers)}, pages 8086--8098,
  2022.

\bibitem[Min et~al.(2022)Min, Lyu, Holtzman, Artetxe, Lewis, Hajishirzi, and
  Zettlemoyer]{min2022rethinking}
Sewon Min, Xinxi Lyu, Ari Holtzman, Mikel Artetxe, Mike Lewis, Hannaneh
  Hajishirzi, and Luke Zettlemoyer.
\newblock Rethinking the role of demonstrations: What makes in-context learning
  work?
\newblock \emph{arXiv preprint arXiv:2202.12837}, 2022.

\bibitem[Ott et~al.(2018)Ott, Auli, Grangier, and Ranzato]{ott2018analyzing}
Myle Ott, Michael Auli, David Grangier, and Marc’Aurelio Ranzato.
\newblock Analyzing uncertainty in neural machine translation.
\newblock In \emph{International Conference on Machine Learning}, pages
  3956--3965. PMLR, 2018.

\bibitem[Papangelis et~al.(2021)Papangelis, Gopalakrishnan, Padmakumar, Kim,
  Tur, and Hakkani-Tur]{papangelis-etal-2021-generative}
Alexandros Papangelis, Karthik Gopalakrishnan, Aishwarya Padmakumar, Seokhwan
  Kim, Gokhan Tur, and Dilek Hakkani-Tur.
\newblock Generative conversational networks.
\newblock In \emph{Proceedings of the 22nd Annual Meeting of the Special
  Interest Group on Discourse and Dialogue}, pages 111--120, Singapore and
  Online, July 2021. Association for Computational Linguistics.

\bibitem[Paszke et~al.(2019)Paszke, Gross, Massa, Lerer, Bradbury, Chanan,
  Killeen, Lin, Gimelshein, Antiga, et~al.]{paszke2019pytorch}
Adam Paszke, Sam Gross, Francisco Massa, Adam Lerer, James Bradbury, Gregory
  Chanan, Trevor Killeen, Zeming Lin, Natalia Gimelshein, Luca Antiga, et~al.
\newblock Pytorch: An imperative style, high-performance deep learning library.
\newblock \emph{Advances in neural information processing systems}, 32, 2019.

\bibitem[Pavlopoulos et~al.(2020)Pavlopoulos, Sorensen, Dixon, Thain, and
  Androutsopoulos]{pavlopoulos2020toxicity}
John Pavlopoulos, Jeffrey Sorensen, Lucas Dixon, Nithum Thain, and Ion
  Androutsopoulos.
\newblock Toxicity detection: Does context really matter?
\newblock \emph{arXiv preprint arXiv:2006.00998}, 2020.

\bibitem[Perez et~al.(2021)Perez, Kiela, and Cho]{perez2021true}
Ethan Perez, Douwe Kiela, and Kyunghyun Cho.
\newblock True few-shot learning with language models.
\newblock \emph{Advances in Neural Information Processing Systems}, 34, 2021.

\bibitem[Robinson et~al.(2020)Robinson, Jegelka, and Sra]{robinson2020strength}
Joshua Robinson, Stefanie Jegelka, and Suvrit Sra.
\newblock Strength from weakness: fast learning using weak supervision.
\newblock In \emph{Proceedings of the 37th International Conference on Machine
  Learning}, pages 8127--8136, 2020.

\bibitem[Sahu et~al.(2022)Sahu, Rodriguez, Laradji, Atighehchian, Vazquez, and
  Bahdanau]{sahu2022data}
Gaurav Sahu, Pau Rodriguez, Issam Laradji, Parmida Atighehchian, David Vazquez,
  and Dzmitry Bahdanau.
\newblock Data augmentation for intent classification with off-the-shelf large
  language models.
\newblock In \emph{Proceedings of the 4th Workshop on NLP for Conversational
  AI}, pages 47--57, 2022.

\bibitem[Schick et~al.(2021)Schick, Udupa, and Sch{\"u}tze]{schick2021self}
Timo Schick, Sahana Udupa, and Hinrich Sch{\"u}tze.
\newblock Self-diagnosis and self-debiasing: A proposal for reducing
  corpus-based bias in nlp.
\newblock \emph{Transactions of the Association for Computational Linguistics},
  9:\penalty0 1408--1424, 2021.

\bibitem[Schuster et~al.(2019)Schuster, Gupta, Shah, and
  Lewis]{schuster2019cross}
Sebastian Schuster, Sonal Gupta, Rushin Shah, and Mike Lewis.
\newblock Cross-lingual transfer learning for multilingual task oriented
  dialog.
\newblock In \emph{Proceedings of the 2019 Conference of the North American
  Chapter of the Association for Computational Linguistics: Human Language
  Technologies, Volume 1 (Long and Short Papers)}, pages 3795--3805, 2019.

\bibitem[Seidenfeld(1986)]{seidenfeld1986entropy}
Teddy Seidenfeld.
\newblock Entropy and uncertainty.
\newblock \emph{Philosophy of Science}, 53\penalty0 (4):\penalty0 467--491,
  1986.

\bibitem[Soltan et~al.(2022)Soltan, Ananthakrishnan, FitzGerald, Gupta, Hamza,
  Khan, Peris, Rawls, Rosenbaum, Rumshisky, et~al.]{soltan2022alexatm}
Saleh Soltan, Shankar Ananthakrishnan, Jack FitzGerald, Rahul Gupta, Wael
  Hamza, Haidar Khan, Charith Peris, Stephen Rawls, Andy Rosenbaum, Anna
  Rumshisky, et~al.
\newblock Alexatm 20b: Few-shot learning using a large-scale multilingual
  seq2seq model.
\newblock \emph{arXiv preprint arXiv:2208.01448}, 2022.

\bibitem[Wang(2021)]{mesh-transformer-jax}
Ben Wang.
\newblock {Mesh-Transformer-JAX: Model-Parallel Implementation of Transformer
  Language Model with JAX}.
\newblock \url{https://github.com/kingoflolz/mesh-transformer-jax}, May 2021.

\bibitem[Wang and Komatsuzaki(2021)]{gpt-j}
Ben Wang and Aran Komatsuzaki.
\newblock {GPT-J-6B: A 6 Billion Parameter Autoregressive Language Model}.
\newblock \url{https://github.com/kingoflolz/mesh-transformer-jax}, May 2021.

\bibitem[Wang(2008)]{wang2008probability}
Qiuping~A Wang.
\newblock Probability distribution and entropy as a measure of uncertainty.
\newblock \emph{Journal of Physics A: Mathematical and Theoretical},
  41\penalty0 (6):\penalty0 065004, 2008.

\bibitem[Wang et~al.(2020)Wang, Zhang, Ma, Wang, and
  Xiao]{wang2020contextualized}
Yan Wang, Jiayu Zhang, Jun Ma, Shaojun Wang, and Jing Xiao.
\newblock Contextualized emotion recognition in conversation as sequence
  tagging.
\newblock In \emph{Proceedings of the 21th annual meeting of the special
  interest group on discourse and dialogue}, pages 186--195, 2020.

\bibitem[Wei and Zou(2019)]{wei-zou-2019-eda}
Jason Wei and Kai Zou.
\newblock {EDA}: Easy data augmentation techniques for boosting performance on
  text classification tasks.
\newblock In \emph{Proceedings of the 2019 Conference on Empirical Methods in
  Natural Language Processing and the 9th International Joint Conference on
  Natural Language Processing (EMNLP-IJCNLP)}, pages 6382--6388, Hong Kong,
  China, November 2019. Association for Computational Linguistics.
\newblock \doi{10.18653/v1/D19-1670}.
\newblock URL \url{https://aclanthology.org/D19-1670}.

\bibitem[Wei et~al.(2022)Wei, Wang, Schuurmans, Bosma, Chi, Le, and
  Zhou]{wei2022chain}
Jason Wei, Xuezhi Wang, Dale Schuurmans, Maarten Bosma, Ed~Chi, Quoc Le, and
  Denny Zhou.
\newblock Chain of thought prompting elicits reasoning in large language
  models.
\newblock \emph{arXiv preprint arXiv:2201.11903}, 2022.

\bibitem[Wolf et~al.(2020)Wolf, Debut, Sanh, Chaumond, Delangue, Moi, Cistac,
  Rault, Louf, Funtowicz, et~al.]{wolf2020transformers}
Thomas Wolf, Lysandre Debut, Victor Sanh, Julien Chaumond, Clement Delangue,
  Anthony Moi, Pierric Cistac, Tim Rault, R{\'e}mi Louf, Morgan Funtowicz,
  et~al.
\newblock Transformers: State-of-the-art natural language processing.
\newblock In \emph{Proceedings of the 2020 conference on empirical methods in
  natural language processing: system demonstrations}, pages 38--45, 2020.

\bibitem[Xiao and Wang(2019)]{xiao2019quantifying}
Yijun Xiao and William~Yang Wang.
\newblock Quantifying uncertainties in natural language processing tasks.
\newblock In \emph{Proceedings of the AAAI Conference on Artificial
  Intelligence}, volume~33, pages 7322--7329, 2019.

\bibitem[Xie et~al.(2020)Xie, Dai, Hovy, Luong, and Le]{xie2020unsupervised}
Qizhe Xie, Zihang Dai, Eduard Hovy, Thang Luong, and Quoc Le.
\newblock Unsupervised data augmentation for consistency training.
\newblock \emph{Advances in Neural Information Processing Systems},
  33:\penalty0 6256--6268, 2020.

\bibitem[Xue et~al.(2021)Xue, Constant, Roberts, Kale, Al-Rfou, Siddhant,
  Barua, and Raffel]{xue2021mt5}
Linting Xue, Noah Constant, Adam Roberts, Mihir Kale, Rami Al-Rfou, Aditya
  Siddhant, Aditya Barua, and Colin Raffel.
\newblock mt5: A massively multilingual pre-trained text-to-text transformer.
\newblock In \emph{Proceedings of the 2021 Conference of the North American
  Chapter of the Association for Computational Linguistics: Human Language
  Technologies}, pages 483--498, 2021.

\bibitem[Yang et~al.(2020)Yang, Malaviya, Fernandez, Swayamdipta, Le~Bras,
  Wang, Bhagavatula, Choi, and Downey]{yang-etal-2020-generative}
Yiben Yang, Chaitanya Malaviya, Jared Fernandez, Swabha Swayamdipta, Ronan
  Le~Bras, Ji-Ping Wang, Chandra Bhagavatula, Yejin Choi, and Doug Downey.
\newblock Generative data augmentation for commonsense reasoning.
\newblock In \emph{Findings of the Association for Computational Linguistics:
  EMNLP 2020}, pages 1008--1025. Association for Computational Linguistics,
  2020.

\bibitem[Zhang et~al.(2022)Zhang, Roller, Goyal, Artetxe, Chen, Chen, Dewan,
  Diab, Li, Lin, Mihaylov, Ott, Shleifer, Shuster, Simig, Koura, Sridhar, Wang,
  and Zettlemoyer]{zhang2022opt}
Susan Zhang, Stephen Roller, Naman Goyal, Mikel Artetxe, Moya Chen, Shuohui
  Chen, Christopher Dewan, Mona Diab, Xian Li, Xi~Victoria Lin, Todor Mihaylov,
  Myle Ott, Sam Shleifer, Kurt Shuster, Daniel Simig, Punit~Singh Koura, Anjali
  Sridhar, Tianlu Wang, and Luke Zettlemoyer.
\newblock Opt: Open pre-trained transformer language models, 2022.

\bibitem[Zhao et~al.(2021)Zhao, Wallace, Feng, Klein, and
  Singh]{zhao2021calibrate}
Zihao Zhao, Eric Wallace, Shi Feng, Dan Klein, and Sameer Singh.
\newblock Calibrate before use: Improving few-shot performance of language
  models.
\newblock In \emph{International Conference on Machine Learning}, pages
  12697--12706. PMLR, 2021.

\end{thebibliography}
\clearpage

\appendix
\begin{figure}[t]
    \centering
    \includegraphics[width=0.75\linewidth]{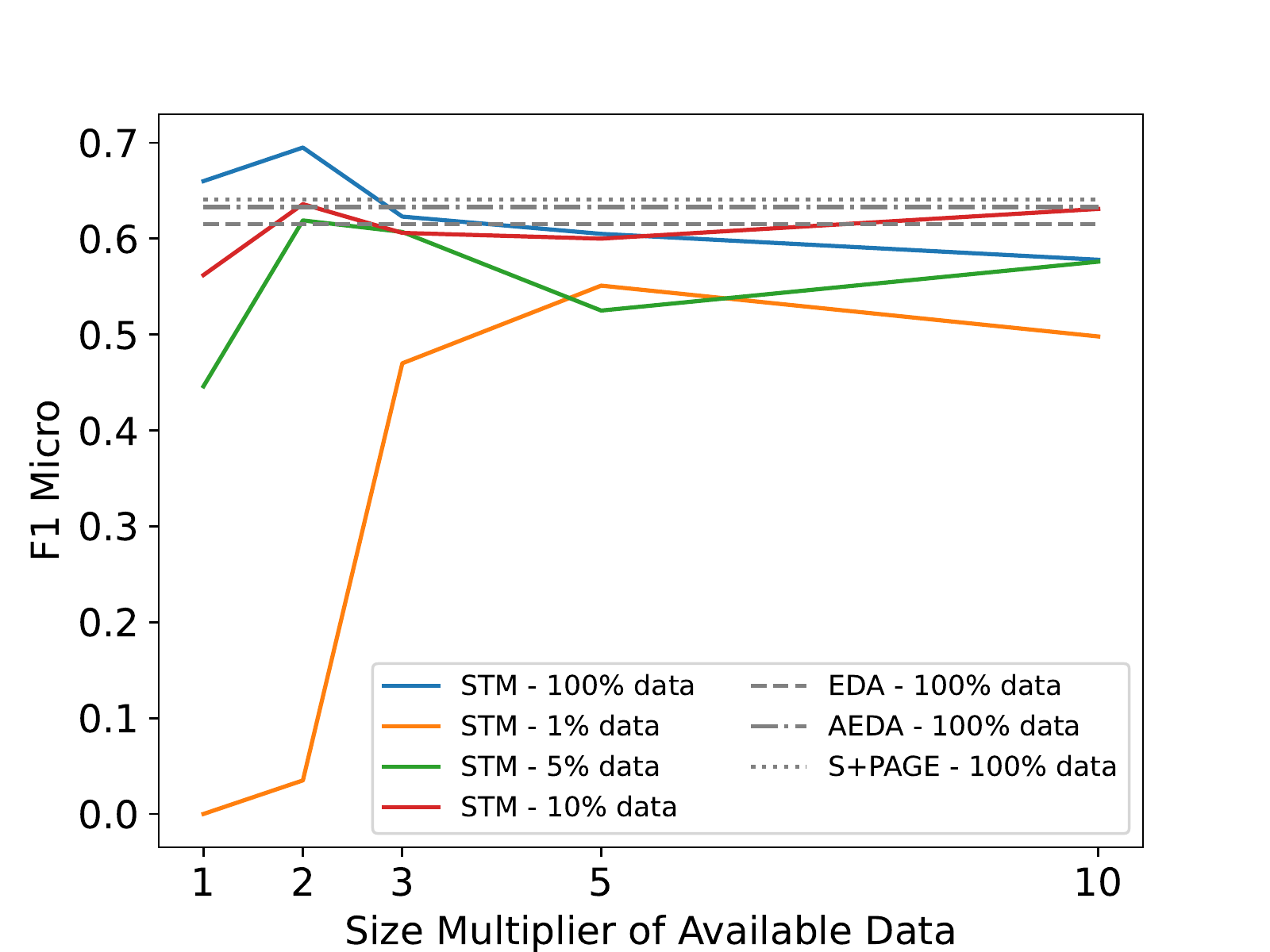}
    \caption{Classification results on the \dyda emotion classification task using Conversation Trajectory Augmentation. Multiplier size represents the dataset size in multiples of the amount of gold data. $n\%$ represents the percentage of the gold data used.}
    \label{fig:emotion_trajectory}
\end{figure}

\begin{figure*}[t]
    \centering
    \includegraphics[width=\linewidth]{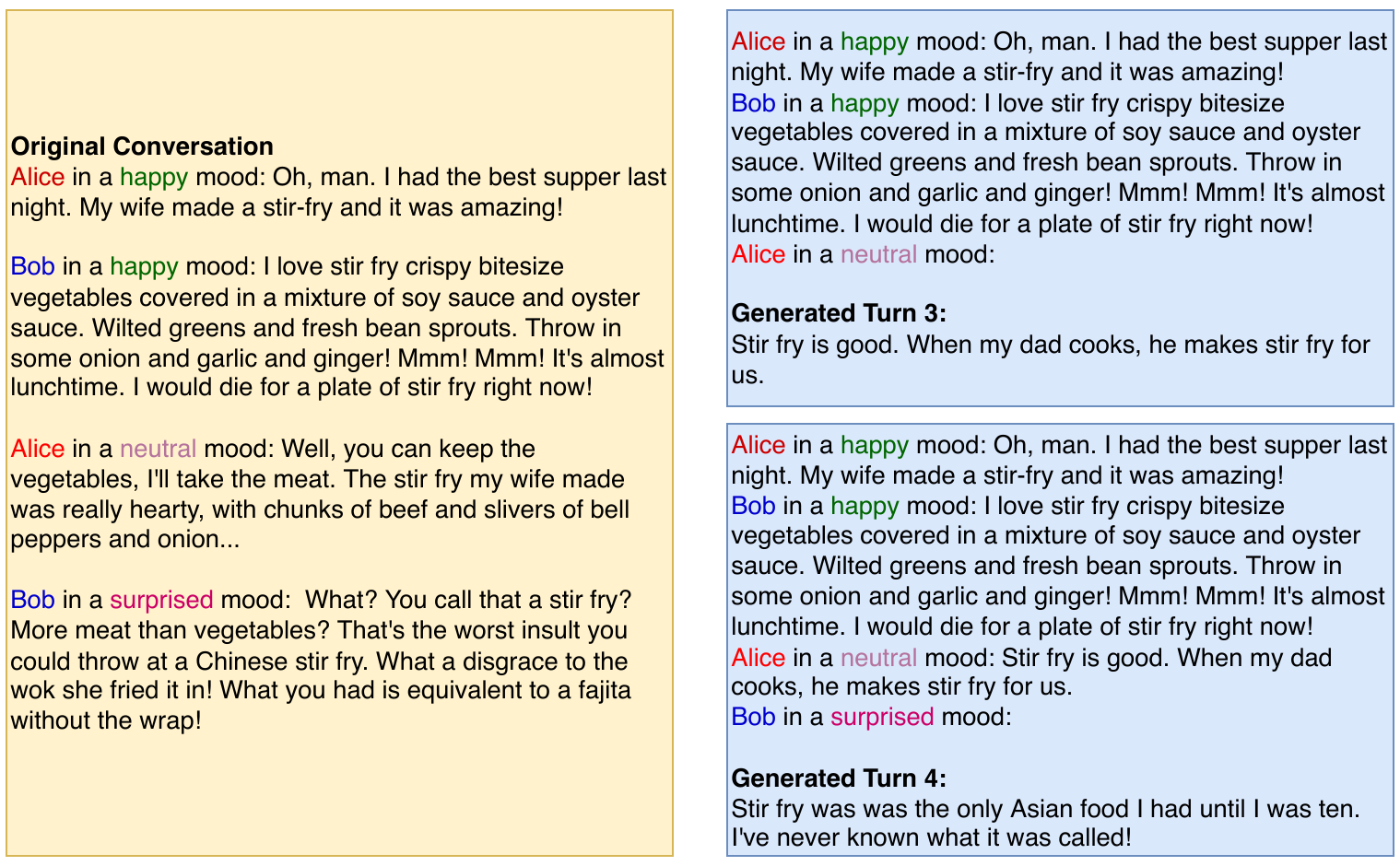}
    \caption{Example prompt using Conversation Trajectory Augmentation for emotion classification using GPT-J. The original conversation (left) has four turns. After allowing each speaker to speak once, GPT-J autoregressively generates the rest of the conversation. Turn 3 (top right) is generated using the first two ground truth turns, and Turn 4 (bottom right) is generated using the first two ground truth turns and the generated third turn. The final candidate conversation is represented by all of the turns in the bottom right box.}
    \label{fig:conversation_trajectory_example}
\end{figure*}
\begin{figure*}[t]
    \centering
    \includegraphics[width=\linewidth]{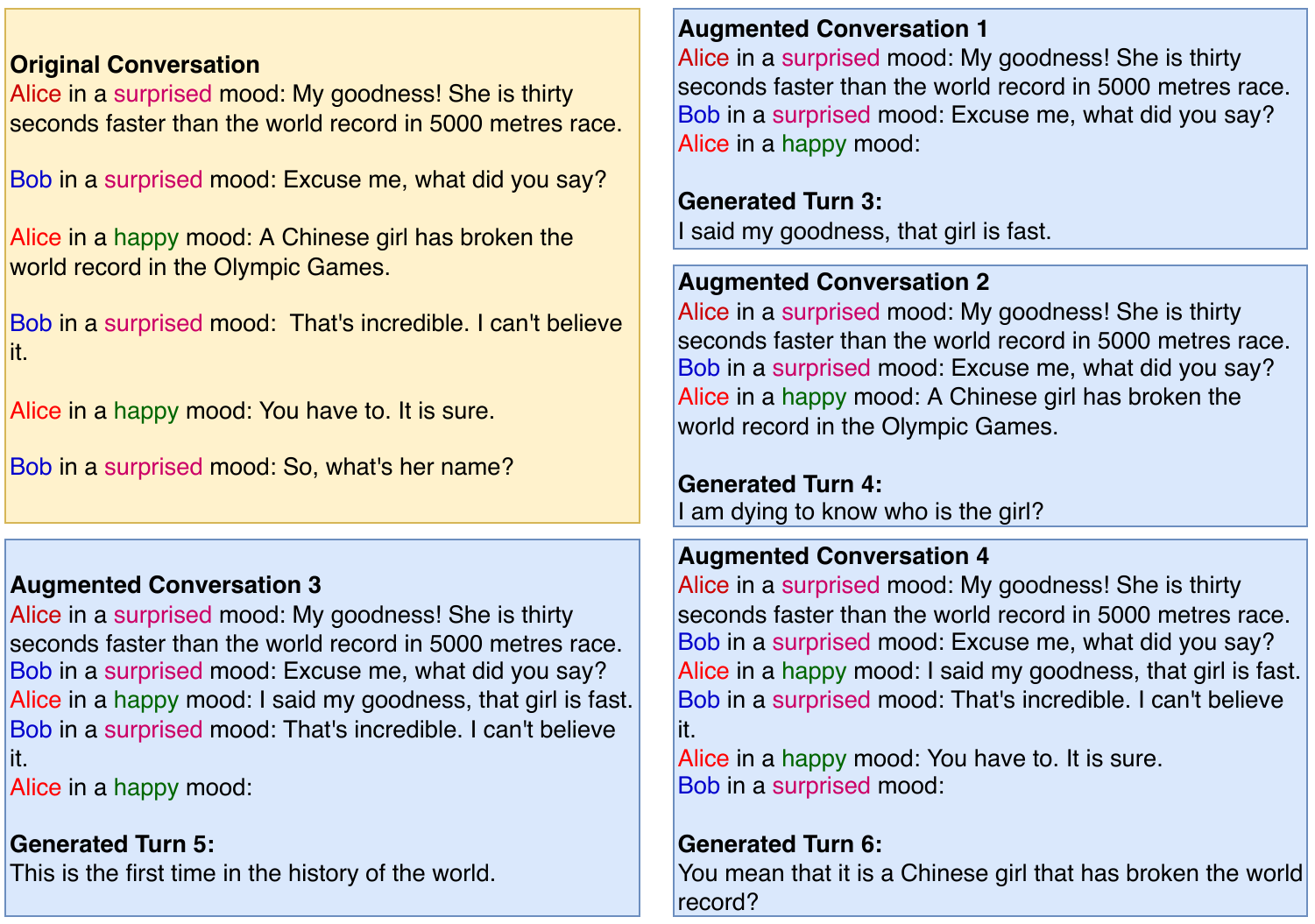}
    \caption{Example prompt using All Turn Augmentation for emotion classification using GPT-J. The original conversation (left) has four turns. After allowing each speaker to speak once, GPT-J generates turn three using the first two ground truth turns (top right), turn four using the first three ground truth turns (middle right), turn five using the first four ground truth turns (bottom left), and turn six using the first five ground truth turns (bottom right). Each of the blue boxes represents a new conversation with the corresponding generated turn as the new endpoint.}
    \label{fig:all_turn_example}
\end{figure*}
\begin{figure}
    \centering
    \includegraphics{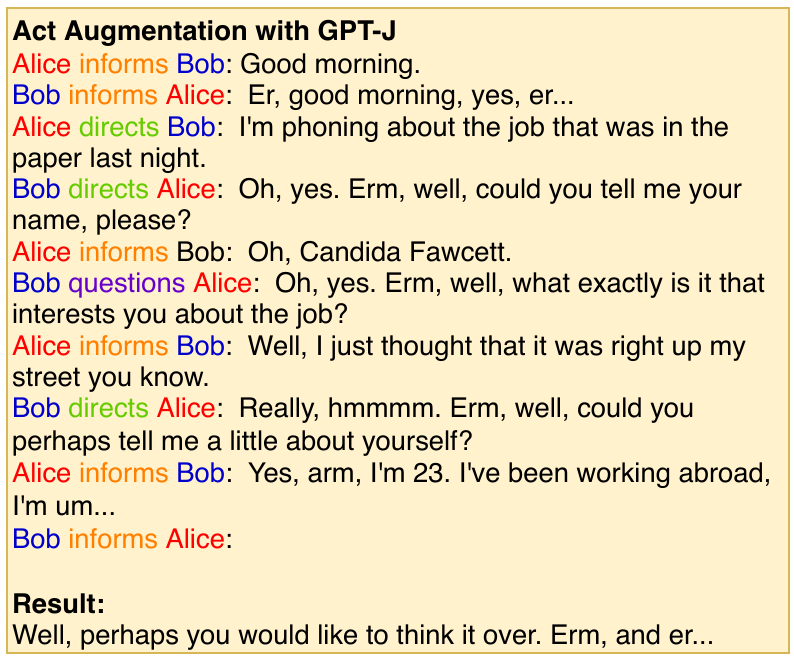}
    \caption{Example prompt using Last Turn Augmentation for act augmentation using GPT-J. The original conversation had nine turns, and the augmented conversation is the same except with the last turn replaced with a generated utterance. GPT-J learns to use some of the speaker's tendencies, such as using the term ``Erm.''}
    \label{fig:act_prompt}
\end{figure}
\section{Ethical Considerations}
\label{ethics}
\paragraph{Pre-trained Language Model Biases} In this work, we directly use only two datasets: \dyda and \fbtodnospace. However, large pre-trained language models like GPT-J have already been pre-trained on massive corpora such as The Pile~\citep{gao2020pile}, which is a webcrawl of much of the internet. While this promotes diverse language, this also provides no guarantee over the types of content that the model is capable of producing. It is possible that the model could generate negative, hurtful, or offensive content~\citep{gao2020pile,christian2020alignment}. Prompting methods as the ones proposed in this work fundamentally rely on large pre-trained language models. As such, it is possible that this hurtful content could leak into augmented datasets, if left unchecked. While in this work we only discuss dialogue understanding tasks, this may present a bigger issue in dialogue generation where hurtful utterances may actually be relayed to users. There are several works that attempt to mitigate these issues, for example \cite{lauscher2021sustainable, schick2021self, barikeri2021redditbias, faal2022reward, pavlopoulos2020toxicity}.

\paragraph{Data Biases} In a similar vein, every dataset, including \dyda and \fbtod has its own biases. While introducing language model output from another language manifold (i.e., a pre-trained language model's training corpora) will help to lessen some of the biases present in unaugmented datasets, generation conditioned solely on existing dialogue contexts may continue to reinforce some of these existing biases.

\paragraph{Reproducibility} 
Upon acceptance, we plan to release all of the augmentation code, as well as the seed data (i.e. each few-shot setting) used in all experiments. 

\section{Limitations}
In this work, we consider two different dialogue contexts --- social chit-chat in \dyda and multi-lingual task-oriented dialogue in \fbtodfullnospace. Due to computational constraints, it is difficult to consider several dialogue contexts with all of our experimental settings. However, we believe that we present a set of experiments representative of the scope of our method's generalizability to different dialogue understanding tasks and integratability with different pre-trained language models.

\paragraph{Prompt Selection} 
Much recent research focuses on how to design prompts to optimize performance on a variety of different tasks~\cite{liu2021pre}. In this work, we did not apply such prompt engineering methods. Instead, we focused on the conversational nature of prompting for augmentation in dialogue tasks. Our prompt primarily consists of dialogue context. Design decisions included the use of the names ``Alice'' and ``Bob,'' as well as the choice to encode instance labels in natural language form. We did not formally evaluate these decisions, but we subjectively saw that the use of other names can yield similar performance. We also noticed that using names generally provided better results than generic speaker tags such as ``Speaker 1'' and ``Speaker 2.''

\section{Datasets} 
\label{sec:datasets}
\dyda \citep{li2017dailydialog} is a high-quality open-domain conversation dataset. The official \dyda training set contains 11,118 dialogues, while the validation and test sets each have 1,000 dialogues. Each conversation is annotated with a topic label and has on average eight turns. Additionally, each turn is annotated with a dialogue act and an emotion label. 

Facebook Multilingual Task-Oriented Dialogue (\textsc{FBTOD}; \citet{schuster2019cross}) is a dataset which contains single-turn task-oriented utterances in English, Spanish, and Thai. Each utterance is annotated with one of twelve intent labels.

\section{\dyda Experiments}
In order to measure and isolate the effect of prompting methods, we hold many of the experimental settings fixed. For data generation with GPT-J, we use top-$p$ sampling with $p=0.92$. The resulting generated data is parsed from the decoded language model outputs. All of our downstream classification experiments are performed using STM. In order to isolate the effects of augmentation and due to computational limitations, we fix the tunable STM parameters with an initial learning rate of $.0001$, $2$ recurrent layers and a $50\%$ dropout layer. We let STM fine-tune for up to 100 epochs; we use early stopping with a patience of 10 epochs.

All experiments are implemented with \texttt{PyTorch}~\citep{paszke2019pytorch} and HuggingFace's \texttt{Transformers}~\citep{wolf2020transformers}, and run on AWS {\it p3.16xlarge} EC2 instances.

\subsection{Conversation Trajectory Augmentation}
In Figure~\ref{fig:emotion_trajectory}, we see that we are able to reach state-of-the-art performance on the emotion classification task using conversation trajectory augmentation as well. Augmented full-shot STM reaches an F1 score of 0.695. In all except the $1\%$ ground-truth setting, Conversation Trajectory Augmentation underperforms Last Turn Augmentation. This is likely due to the fact that Conversation Trajectory Augmentation departs from the ground-truth data distribution, in contrast to Last Turn Augmentation as displayed in Figure~\ref{fig:manifold_comparison}. This is not necessarily a downside of generating new trajectories - it just indicates that it does not perform as strongly on this particular classification task. By definition, training a model on data that closely matches the testing distribution will be advantageous during evaluation. However, beyond classification tasks, it is possible that this diverse data distribution will be more favorable, e.g., in response and dialogue generation, where one would want to see more diverse responses.

\subsection{All-Turn Augmentation}
\begin{table}[]
\centering
\begin{tabular}{lll}
Size & Augmented Mult. & F1 \\ \hline
1\%  & 5.7x             & 0.459   \\
5\%  & 6.2x               & 0.463  \\
10\% & 6.1x               & 0.610  
\end{tabular}
\caption{Resulting augmentation sizes and classification F1 Micro using All-Turn Augmentation on the \dyda emotion classification task.}
\label{allturn_classification}
\end{table}
As displayed in Table~\ref{allturn_classification},  All-Turn Augmentation generally does not outperform either Conversation Trajectory Augmentation nor Last-Turn Augmentation. Moreover, we see that the resulting augmented sizes are roughly six times that of the original data. The closest comparable augmented size used with Last-Turn and Conversation Trajectory Augmentation is a multiplier of five, but at that multiple, either Last Turn or Conversation Trajectory Augmentation yields the strongest performance for each data setting.
\section{Example Prompts}
\label{sec:example_prompts}
We present several examples of prompts corresponding to the different methods of conversation augmentation. 

Figure~\ref{fig:last_turn_example} shows an example of Last Turn Augmentation and demonstrates that by omitting the last turn of the original conversation, we can form a prefix prompt that allows GPT-J to generate an utterance according to a prescribed emotion, whether it be the ground truth emotion from the training data or a user-defined emotion. The ground truth turns given as context combined with the generated utterance constitute a new augmented conversation.

Figure~\ref{fig:conversation_trajectory_example} shows an example of Conversation Trajectory Augmentation. In order to set the context for each speaker, we always include the first ground truth turn for each speaker in the context. For each subsequent turn, we autoregressively let the generation model generate the utterance by feeding in the previously generated utterance. This process continues until the number of turns in the new augmented conversation reaches the length of the original ground truth conversation.

We show an example of All Turn Augmentation in Figure~\ref{fig:all_turn_example}. We again include the first ground truth turn for each speaker in the conversation. However, instead of autoregressively feeding in previously generated utterances, the context of the generated turn $i$ is always the ground truth turns $1$ through $i-1$. Since the generated turn $i$ is not guaranteed to form a coherent conversation when with ground truth turns $i+1$ through $n$ (where $n$ is the length of the ground truth conversation), we consider each generated turn to be an endpoint for a new conversation.

Finally, we show an example using Last Turn Augmentation while prescribing a \dyda act label in Figure \ref{fig:act_prompt}. We see that GPT-J correctly generates an utterance that follows the act ``inform,'' and is even able to pick up on some of the speaker's tendencies (e.g., using the onomatopoeia ``Erm'').
\begin{figure}[t]
    \centering
    \includegraphics{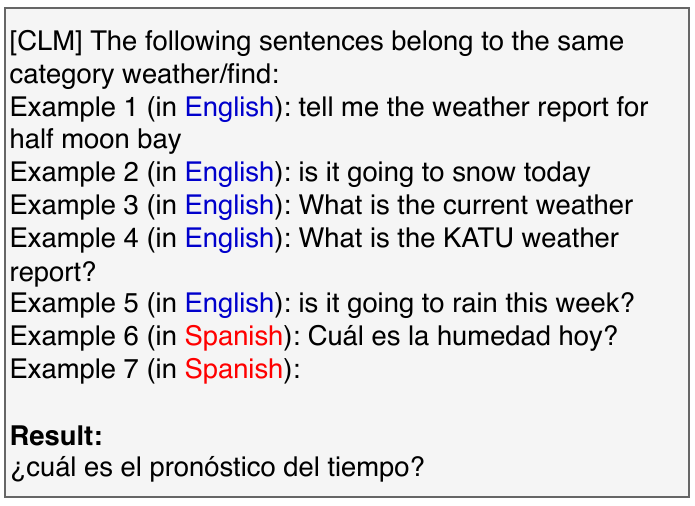}
    \caption{Cross-lingual prompt between Spanish and English for \fbtod intent detection. [CLM] is a token reserved for the model that is required for in-context prompts.}
    \label{fig:intent_prompt}
\end{figure}
\section{Cross-lingual Augmentation Experimental Setup}
\label{sec:crosslingualsetup}
The original \fbtod dataset contains an English dataset with 30,521 training instances, 4,181 evaluation instances, and 8,621 testing instances. It also contains a Spanish dataset with 3,617 training instances, 1,983 evaluation instances, and 3,043 testing instances. For both languages, we created versions of the training set that had $1\%$, $5\%$, and $10\%$ of the original number of training instances. For each percentage, we randomly sampled that proportion of training examples for each intent label. In cases where that proportion would result in a number smaller than $1.0$, we ensured that there would be at least one training instance.

For augmentation, we use the prompt given in Figure~\ref{fig:intent_prompt}. We perform beam search, taking up to three sequences as augmented data and rejecting any duplicate examples. For classification, we fine-tune XLMRoBERTa with a maximum sequence length of 128, 80 training epochs, and an initial learning rate of 5e-5. 
\section{Entropy-based Weak Filtering}
\label{sec:entropy_discussion}
We hypothesized that two categories of synthetic data could improve to eventual task performance. The first is simply \textit{correct} data --- synthetic training instances which match the intended instance label. However, we determine this correctness using an imperfect classifier. Thus, the second category is \textit{``hard to learn''} data. This follows from early work~\citep{guo2004boosting} finding that classification performance can be improved by focusing on augmenting datasets with ``hard to learn'' examples. However \citet{guo2004boosting} identify such difficult examples post hoc through classifier performance. Solely evaluating classifier performance for each example is prohibitively expensive for large models on large datasets. Thus, in our case, we make our decisions based on uncertainty. Out of those instances that the classifier states are incorrect, we do not filter out the ones for which the classification is made with high uncertainty, hypothesizing that uncertainty is a strong proxy for data points' learning difficulty, which may be more useful in training the next iteration of the classifier.

There are several ways to consider how to quantify uncertainty. 
\citet{ott2018analyzing} quantified uncertainty for neural machine translation post hoc by comparing generation methods' sequence-level probability mass coverage of ground-truth translations. While these methods may be appropriate for translation or even response generation-related tasks, there is no reference point for what the ideal augmented data is. 

Other approaches have looked specifically at quantifying dataset uncertainty using Bayesian Neural Networks~\citep{xiao2019quantifying, chen2020uncertainty}. However, dataset level approaches are not necessarily appropriate for augmentation, where one needs to make inferences about data quality at the instance level. Moreover, approaches requiring Bayesian Neural Networks do not achieve state-of-the-art performance on the \dyda classification tasks.
 
On the other hand, entropy maximization has long been used as an information theoretic technique for maximizing uncertainty~\citep{seidenfeld1986entropy,wang2008probability}. Even in natural language processing, \citet{csaky2019improving} performs data filtering using entropy to improve diversity by specifically removing generic data points. They achieve promising results in dialogue generation, and moreover, entropy computation is directly applicable at the instance level. Therefore, we adopt entropy into our framework to identify uncertain and difficult data points.
\end{document}